\documentclass[11pt]{article}

\pdfoutput=1  

\usepackage[hyperref]{acl2017}
\usepackage{times}
\usepackage{url}
\usepackage{latexsym}

\aclfinalcopy 

\usepackage{tikz}
\usetikzlibrary{positioning,arrows,arrows.meta,patterns}
\tikzset{execute at begin node=\strut}
\definecolor{lila}{RGB}{159,114,207}
\definecolor{orange}{RGB}{230,130,50}
\colorlet{plila}{lila!60!white}
\colorlet{porange}{orange!60!white}
\definecolor{grau}{RGB}{192,192,192}

\tikzstyle{default}=[on grid, node distance=21mm and 25mm, x=25mm, y=21mm, style=thick, inner sep=0pt]
\tikzset{execute at begin node=\strut}
\tikzset{
  center coordinate/.style={
    execute at end picture={
      \path ([rotate around={180:#1}]perpendicular cs: horizontal line through={#1},
                                  vertical line through={(current bounding box.east)})
            ([rotate around={180:#1}]perpendicular cs: horizontal line through={#1},
                                  vertical line through={(current bounding box.west)});}}}
\tikzstyle{ent}=[circle, draw, minimum size=10mm, fill=porange]
\tikzstyle{val}=[circle, draw, minimum size=10mm, fill=plila]
\tikzstyle{tok}=[circle, draw, minimum size=10mm, fill=grau]
\tikzstyle{comp}=[rectangle, draw, minimum size=2.1mm, inner sep=-2mm, font=\tiny, fill=orange]
\tikzstyle{sig}=[rectangle, draw, minimum size=2.1mm, inner sep=-2mm, font=\tiny, fill=lila]
\tikzstyle{link1}=[out=106, in=74, looseness=.5]
\tikzstyle{link2}=[out=102, in=78, looseness=.6]
\tikzstyle{link3}=[out=98, in=82, looseness=.69]
\tikzstyle{link4}=[out=94, in=86, looseness=.77]
\tikzstyle{link5}=[out=90, in=90, looseness=.84]
\tikzstyle{dir}=[-{Straight Barb[angle=60:2.2mm]}, shorten > =0.2mm]
\tikzstyle{dircomp}=[{Straight Barb[angle=90:2.2mm]}-, shorten < =0.2mm]
\tikzstyle{dirextra}=[shorten < =0.5mm]
\tikzstyle{inf}=[densely dotted, {Straight Barb[angle=60:2.2mm]}-, shorten < =0.2mm]
\newlength{\compsep}
\usepackage{amssymb}
\usepackage{amsmath}

\usepackage{tabularx} 
\newcolumntype{C}{>{\centering\arraybackslash}X} 

\usepackage{verbatim}
\usepackage{calc}

\usepackage[nameinlink,capitalize]{cleveref}
\crefname{equation}{}{}
\crefformat{section}{#2\S#1#3}
\crefformat{subsection}{#2\S#1#3}

\title{Variational Inference for Logical Inference}

\author{Guy Emerson and Ann Copestake \\
  Computer Laboratory \\
  University of Cambridge \\
  {\tt \{gete2,aac10\}@cam.ac.uk} \\
}

\date{}

\begin{document}
\maketitle
\begin{abstract}
Functional Distributional Semantics
is a framework that aims to learn, from text,
semantic representations which can be interpreted in terms of truth.
Here we make two contributions to this framework.
The first is to show how a type of logical inference can be performed
by evaluating conditional probabilities.
The second is to make these calculations tractable
by means of a variational approximation.
This approximation also enables faster convergence during training,
allowing us to close the gap with state-of-the-art vector space models
when evaluating on semantic similarity.
We demonstrate promising performance on two tasks.
\end{abstract}

\section{Introduction and Background}

Standard approaches to distributional semantics
represent meanings as vectors,
whether this is done using the more traditional \textit{count} vectors
\cite{turney2010vector},
or using \textit{embedding} vectors trained with a neural network
\cite{mikolov2013vector}.
While vector space models have advanced the state of the art in many tasks,
they raise questions when it comes to representing larger phrases.
Ideally, we would like to learn representations
that naturally have logical interpretations.

There have been several attempts to incorporate vectors into logical representations,
and while we do not have space for a full literature review here,
we will mention two prominent lines of research.
\newcite{coecke2010tensor} and \newcite{baroni2014tensor}
propose a tensor-based approach,
where vectors are combined according to argument structure.
However, this leaves open the question of how to perform logical inference,
as vector spaces do not provide a natural notion of entailment.
Indeed, \newcite{grefenstette2013tensor} proved that
quantifiers cannot be expressed using tensor calculus.
\newcite{garrette2011logic} and \newcite{beltagy2016logic}
incorporate a vector space model into a Markov Logic Network,
in the form of weighted inference rules
(the truth of one predicate implying the truth of another).
This approach requires existing vectors,
and assumes we can interpret similarity in terms of inference.

In contrast to the above,
\newcite{emerson2016} (henceforth E\&C)
introduced the framework of Functional Distributional Semantics,
which represents the meaning of a predicate not as a vector, but as a \textit{function}.

To define these functions,
we assume a semantic space~$\mathcal{X}$,
each point representing the features of a possible individual.
We refer to points in~$\mathcal{X}$ as `pixies',
intuitively `pixels' of the space,
to make clear they are not individuals --
two individuals may be represented by the same pixie.
Further discussion of model theory will be given in forthcoming work
\citep{emerson-forth} (henceforth E\&C-forth).
We take~$\mathcal{X}$ to be a vector space,
each dimension intuitively representing a feature.

A semantic function maps from the space~$\mathcal{X}$ to the range~$[0,1]$.
This can be interpreted both in the machine-learning sense of a classifier,
and in the logical sense of a truth-conditional function.\footnote{%
  We take a probabilistic approach,
  where a predicate has a probability of truth for any pixie.
  We believe this is a strength of the model,
  as it can model fuzzy boundaries of concepts.
  However, we could also use semantic functions in a more traditional logic,
  by assigning truth when the function's value is above~$0.5$, and falsehood otherwise.
  This is equivalent to turning the probabilistic classifier into a `hard' classifier.
}
In the machine learning view,
a semantic function is a probabilistic classifier for a binary classification task~--
each input $x\in\mathcal{X}$ is either an instance of the predicate's class, or it is not.
In the logical view,
a semantic function specifies what features a pixie needs to have
in order for the predicate to be true of it~--
that is, the predicate's truth conditions.

This is related to probabilistic type judgements in the framework of
Probabilistic Type Theory with Records (TTR) \cite{cooper2005type,cooper2015prob}.
Working within TTR,
\newcite{larsson2013classifier} argues in favour of
representing perceptual concepts as classifiers of perceptual input.
While TTR represents situations in terms of situation types,
a semantic function model defines a semantic space without reference to any types or predicates.

\newcite{schlangen2016classifier}
take a similar view, representing meanings as image classifiers.
\newcite{zarriess2017classifier}
use a distributional model to help train such classifiers,
but do not directly learn logical representations from distributional data.

Our approach to logical inference is related to the work of
\newcite{bergmair2010fuzzy} and \newcite{clarke2015stochastic},
who use fuzzy truth values and probabilistic truth values, respectively.
However, neither incorporate distributional data.

In contrast to the above,
a semantic function model can be trained on a parsed corpus.
By defining a generative model,
we can apply unsupervised learning:
optimising the model parameters
to maximise the probability of generating the corpus data.

\begin{figure}
\centering

\begin{tikzpicture}[default, center coordinate=(y)]
\tikzstyle{word}=[inner sep=1mm]

\node[word] (x) {dog};
\node[word, right=of x] (y) {chase};
\node[word, right=of y] (z) {cat};

\path (y.center) -- (z.center) node[midway, above] {\textsc{arg2}} ;
\path (y.center) -- (x.center) node[midway, above] {\textsc{arg1}} ;
\draw[dir] (y) edge (z) edge (x);

\end{tikzpicture}
\vspace*{-3mm}

\caption{A simplified DMRS graph,
illustrating the type of data observed during training.
This graph would correspond to sentences like
\textit{The dog chased the cat}, or
\textit{Cats are chased by dogs}.
}
\label{fig:dmrs}

\end{figure}
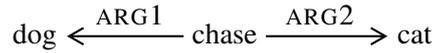

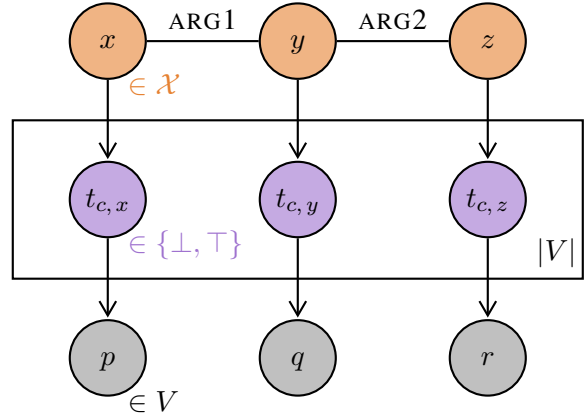
\begin{figure}
\centering

\begin{tikzpicture}[default]


\node[ent] (y) {$y$} ;
\node[ent, right=of y] (z) {$z$} ;
\node[ent, left=of y] (x) {$x$} ;

\draw (y) -- (z) node[midway, above] {\textsc{arg2}} ;
\draw (y) -- (x) node[midway, above] {\textsc{arg1}} ;

\node[below right=3.5mm and 2.5mm of x, anchor=north west] {\textcolor{orange}{$\in \mathcal{X}$}} ;


\draw (-1.5,-0.5) rectangle (1.5,-1.5) ;

\node[val, below=of x] (tx) {$t_{c,\,x}$} ;
\node[val, below=of y] (ty) {$t_{c,\,y}$} ;
\node[val, below=of z] (tz) {$t_{c,\,z}$} ;

\draw[dir] (x) -- (tx);
\draw[dir] (y) -- (ty);
\draw[dir] (z) -- (tz);

\node[below right=3.5mm and 2.5mm of tx, anchor=north west] {\textcolor{lila}{$\in \left\{\bot,\top\right\}$}} ;
\node[xshift=-2ex, yshift=2ex] at (1.5, -1.5) {$|V|$};


\node[tok, below=of tx] (a) {$p$} ;
\node[tok, below=of ty] (b) {$q$} ;
\node[tok, below=of tz] (c) {$r$} ;

\draw[dir] (tx) -- (a);
\draw[dir] (ty) -- (b);
\draw[dir] (tz) -- (c);

\node[below right=3.5mm and 2.5mm of a, anchor=north west] {$\in V$} ;

\end{tikzpicture}
\vspace*{-3mm}

\caption{A probabilistic graphical model for Functional Distributional Semantics
(E\&C Fig.~3).
Each node denotes a random variable, but only the bottom row is observed.
The plate (middle row) denotes repeating variables across the vocabulary. \linebreak
\textbf{Top row:} latent pixies $x$,~$y$,~and~$z$, lying in a\linebreak semantic space~$\mathcal{X}$.
Their joint distribution is determined by the DMRS links \textsc{arg1} and \textsc{arg2}. \linebreak
\textbf{Middle row:} each predicate~$c$ in the vocabulary~$V$ is probabilistically
true or false for each pixie. \linebreak
\textbf{Bottom row:} for each pixie, we observe exactly one predicate,
probabilistically chosen out of all predicates that are true of the pixie.
}
\label{fig:graph}
\vspace*{-3mm}

\end{figure}

Our model can be trained on a corpus annotated with
Dependency Minimal Recursion Semantics (DMRS)
\cite{copestake2005mrs,copestake2009dmrs}.
This represents the meaning of a sentence as a semantic dependency graph
that has a logical interpretation.
An example DMRS graph is shown in \cref{fig:dmrs}
(ignoring quantifiers, and ignoring properties such number and tense).
Note that, as these are semantic dependencies,
not syntactic dependencies,
active and passive voice sentences can be represented with the same graph.
This dependency graph could be generated by the probabilistic graphical model shown in \cref{fig:graph}.
The generated predicates are at the bottom:
$q$~might correspond to a verb, $p$~its subject, and $r$~its object.
The dependency links (\textsc{arg1} and \textsc{arg2}) are at the top.

Rather than generating the predicates directly,
we assume that each predicate is true of a latent, unobserved pixie.
For example, the DMRS graph for \textit{the dog chased the cat} has three pixies,
corresponding to the dog, the chasing event, and the cat.
We can define a generative model for such sets of pixies
(the top row of \cref{fig:graph}),
assuming each link corresponds to a probabilistic dependence;
intuitively, different kinds of events occur with different kinds of arguments.
In machine learning terms, this forms an undirected graphical model.

We generate the predicates in three stages (from top to bottom in \cref{fig:graph}).
First, we generate a set of pixies,
with DMRS links specifying probabilistic dependence.
Second, we use the semantic functions for all predicates
to generate a truth value for each predicate applied to each pixie.
Third, for each pixie, we generate a single predicate out of all true predicates.
The separation of pixies and truth values
gives us a connection with logical inference,
as we will see in~\cref{sec:log}.

For a DMRS graph with a different structure,
we can define a similar graphical model.
For example, for an intransitive sentence,
with just a verb and its subject,
we can remove the right-hand column of \cref{fig:graph}.
The model parameters are shared across all such graphs,
so we can train our model on a heterogenous set of DMRS graphs.

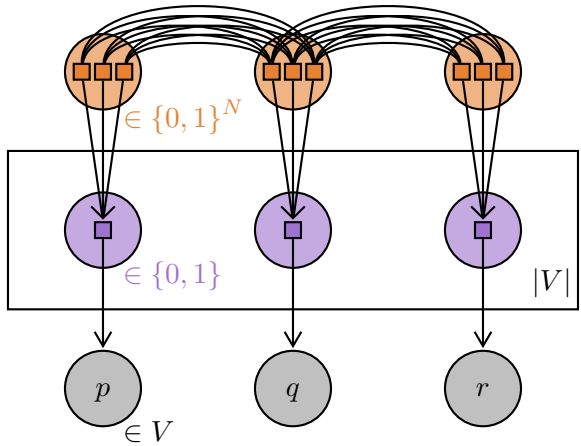
\begin{figure}
\centering

\begin{tikzpicture}[default]


\node[ent] (y) {} ;
\node[ent, right=of y] (z) {} ;
\node[ent, left=of y] (x) {} ;

\node[below right=3.5mm and 2.5mm of x, anchor=north west] {\textcolor{orange}{$\in\left\{0,1\right\}^N$}} ;


\node[comp] (x1) at (x) [xshift=-\compsep] {};
\node[comp] (x2) at (x) {};
\node[comp] (x3) at (x) [xshift=\compsep] {};
\node[comp] (y1) at (y) [xshift=-\compsep] {} edge[link3] (x1) edge[link2] (x2) edge[link1] (x3);
\node[comp] (y2) at (y) {} edge[link4] (x1) edge[link3] (x2) edge[link2] (x3);
\node[comp] (y3) at (y) [xshift=\compsep] {} edge[link5] (x1) edge[link4] (x2) edge[link3] (x3);
\node[comp] (z1) at (z) [xshift=-\compsep] {} edge[link3] (y1) edge[link2] (y2) edge[link1] (y3);
\node[comp] (z2) at (z) {} edge[link4] (y1) edge[link3] (y2) edge[link2] (y3);
\node[comp] (z3) at (z) [xshift=\compsep] {} edge[link5] (y1) edge[link4] (y2) edge[link3] (y3);


\draw (-1.5,-0.5) rectangle (1.5,-1.5) ;

\node[val, below=of x] (tx) {} ;
\node[val, below=of y] (ty) {} ;
\node[val, below=of z] (tz) {} ;

\node[below right=3.5mm and 2.5mm of tx, anchor=north west] {\textcolor{lila}{$\in\left\{0,1\right\}$}} ;
\node[xshift=-2ex, yshift=2ex] at (1.5, -1.5) {$|V|$};


\node[sig] (sx) at (tx) {} ;
\node[sig] (sy) at (ty) {} ;
\node[sig] (sz) at (tz) {} ;

\draw (sx.north) edge[dirextra] (x1) edge[dircomp] (x2) edge[dirextra] (x3);
\draw (sy.north) edge[dirextra] (y1) edge[dircomp] (y2) edge[dirextra] (y3);
\draw (sz.north) edge[dirextra] (z1) edge[dircomp] (z2) edge[dirextra] (z3);


\node[tok, below=of tx] (a) {$p$} ;
\node[tok, below=of ty] (b) {$q$} ;
\node[tok, below=of tz] (c) {$r$} ;

\draw[dir] (sx) -- (a);
\draw[dir] (sy) -- (b);
\draw[dir] (sz) -- (c);

\node[below right=3.5mm and 2.5mm of a, anchor=north west] {$\in V$} ;

\end{tikzpicture}
\vspace*{-3mm}

\caption{Implementation of the model in \cref{fig:graph}.\linebreak
\textbf{Top row:} pixies are binary-valued vectors,
forming a CaRBM.
For each link,
connections between the dimensions of the two pixies
determine how likely they are to be active at the same time.\linebreak
\textbf{Middle row:} each semantic function is a one-layer feedforward network,
with a single output interpreted as the probability of truth. \linebreak
\textbf{Bottom row:} for each pixie, we generate exactly one predicate,
as in \cref{fig:graph}.
}
\label{fig:imp}
\vspace*{-3mm}

\end{figure}

We follow E\&C, and implement this model as shown in \cref{fig:imp}.
The semantic space~$\mathcal{X}$ consists of sparse binary-valued vectors,
where a small fixed number of dimensions are~$1$, and the rest~$0$.
Intuitively, each dimension is a `feature' of a pixie,
and only a small number are present.
The joint distribution over pixies is given by a
Cardinality Restricted Boltzmann Machine (CaRBM) \cite{swersky2012carbm}.
The semantic functions are one-layer feedforward networks,
with a sigmoid activation so the output is in the range~$[0,1]$.
The probability of generating a predicate (bottom row)
is weighted by the observed frequency of the predicate.

\section{Theoretical Contributions}
\label{sec:theory}

\subsection{Logical Inference}
\label{sec:log}

The model in \cref{fig:graph} contains, in the middle row,
a node for the truth of each predicate for each pixie.
Using these nodes, we can convert certain logical propositions
into statements about probabilities.

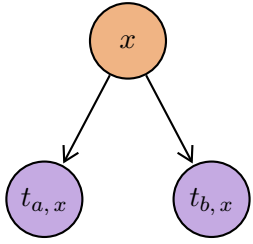
\begin{figure}
\centering

\begin{tikzpicture}[default]

\node[ent] (x) {$x$} ;
\node[val, below=of x, xshift=-11mm] (t1) {$t_{a,\,x}$} ;
\node[val, below=of x, xshift=11mm] (t2) {$t_{b,\,x}$} ;
\draw (x) edge[dir] (t1) edge[dir] (t2) ;

\end{tikzpicture}
\vspace*{-3mm}

\caption{Logical inference for a single pixie~$x$,
and two predicates $a$ and~$b$.
We have a joint distribution between~$x$
and the truth values $t_{a,x}$ and~$t_{b,x}$,
which lets us consider logical inferences in terms of conditional probabilities,
such as $P(t_{b,x}|t_{a,x})$, the probability of $b$ being true, given that $a$ is true.
}
\label{fig:log-one}

\end{figure}

For example, we might be interested in whether one predicate implies another.
For simplicity, consider a single pixie~$x$, as shown in \cref{fig:log-one}.
Then, the logical proposition ${\forall x\in\mathcal{X},\; a(x) \Rightarrow b(x)}$\linebreak
is equivalent\footnote{%
  More precisely, the equivalence requires the logic to have `existential import':
  \textit{Every A} implies that some \textit{A} exists.
  This follows from the definition of conditional probability
  $P(B|A)=P(A\wedge B)/P(A)$, only defined if $P(A)\neq 0$
} 
to the statement ${P(t_{b,x}|t_{a,x}) = 1}$.\linebreak
Intuitively, conditioning on~$t_{a,x}$ means restricting to those pixies~$x$
for which the predicate~$a$ is true.
If the probability of $t_{b,x}$ being true is~$1$,
then the predicate~$b$ is true for all of those~$x$.
Similarly, ${\exists x\in\mathcal{X},\; a(x) \land b(x)}$
is equivalent to ${P(t_{b,x}|t_{a,x}) > 0}$.
Furthermore, classical rules of inference hold under this equivalence.
For example, from ${P(t_{b,x}|t_{a,x}) = 1}$ and ${P(t_{c,x}|t_{b,x}) = 1}$,
we can deduce that ${P(t_{c,x}|t_{a,x}) = 1}$.
This is precisely the classical Barbara syllogism.
A proof is given in \cref{append:equiv}.

In practice, when training on distributional data,
the conditional probability ${P(t_{b,x}|t_{a,x})}$ will never be exactly $0$ or~$1$,
because the model only implements soft constraints.
Nonetheless, this quantity can be very informative:
if it is $0.999$, then we know that if $a(x)$ is true,
it is \textit{almost always} the case that $b(x)$ is also true.
So, it represents the degree to which $a$ implies~$b$, in an intuitive sense.

Separate from this notion of inference,
we can also consider the similarity of two latent pixies --
if $a$ is true of~$x$, and $b$ is true of~$y$,
how many features do $x$ and $y$ share?
If $a$ and~$b$ are antonyms,
the truth of one will not imply the truth of the other,
but the pixies may share many features.

As \newcite{copestake2012ideal} note,
distinguishing synonyms and antonyms
requires checking whether expressions are mutually exclusive.
We do not have access to such information in our training data,
and such cases are inconsistently annotated in our test data (see~\cref{sec:sim}).
Nonetheless, the model allows us to make such a distinction,
which is an advantage over vector space models.
Exploiting this distinction (perhaps by using coreference information)
would be a task for future work.

To calculate ${P(t_{b,x}|t_{a,x})}$, we must marginalise out~$x$,
because the model actually defines the joint probability ${P(x,\,t_{b,x},\,t_{a,x})}$.
This is analogous to removing bound variables
when calculating the truth of quantified expressions in classical logic.
Quantifiers will be discussed further by E\&C-forth.
However, marginalising out~$x$ requires summing over the semantic space~$\mathcal{X}$,
which is intractable when~$\mathcal{X}$ has a large number of dimensions.
In~\cref{sec:var}, we introduce a variational approximation
to make this calculation tractable.

\begin{figure}
\centering
\begin{tikzpicture}[default]


\node[ent] (y) {$y$} ;
\node[ent, right=of y] (z) {$z$} ;
\node[ent, left=of y] (x) {$x$} ;

\draw (y) -- (z) node[midway, above] {\textsc{arg2}} ;
\draw (y) -- (x) node[midway, above] {\textsc{arg1}} ;


\node[val, below=of x] (tx) {$t_{a,\,x}$} ;
\node[val, below=of y] (ty) {$t_{b,\,y}$} ;
\node[val, below=of z] (tz) {$t_{c,\,z}$} ;
\node[val, above=of x] (t2) {$t_{d,\,x}$} ;

\draw[dir] (x) -- (tx);
\draw[dir] (y) -- (ty);
\draw[dir] (z) -- (tz);
\draw[dir] (x) -- (t2);

\end{tikzpicture}
\vspace*{-3mm}

\caption{Logical inference for three pixies
and four predicates of interest:
we know whether $a$,~$b$,~$c$ are true of $x$,~$y$,~$z$, respectively,
and we would like to infer whether $d$ is true of~$x$.
The distribution for $t_{d,x}$
depends on all the other truth values,
because it is indirectly connected to them via the latent pixies.
}
\label{fig:log-three}

\end{figure}
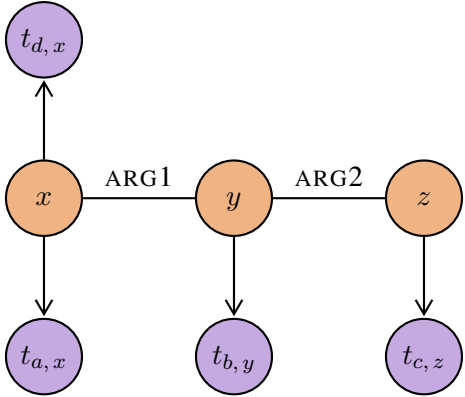

In the general case, there are multiple pixie variables.
This opens up the possibility of inferring what is true of one pixie,
given what is true of another.
For example, we might be interested in what is true of a verb's arguments,
which we could explore with the three-pixie graph in \cref{fig:log-three}.
We can ask questions such as: if the predicate \textit{paint} is true of an event,
what predicates are true of its arguments?
A good model might answer that for the \textsc{arg1} pixie,
\textit{artist} and \textit{person} are likely true,
while \textit{democracy} and \textit{aubergine} are likely false.

Just as with the one-pixie case,
there is an equivalence between logical propositions and statements about probabilities.
For example, ${\exists x,y\in\mathcal{X},\; a(y) \land b(x) \land \textsc{arg1}(y,x)}$
is equivalent to ${P(t_{b,x}|t_{a,y})>0}$.
Note that \textsc{arg1} does not correspond to a random variable --
it is instead represented directly by the structure of the graphical model
(the edges in the top row of \cref{fig:graph}
and the middle row of \cref{fig:log-three}).
As before, this conditional probability is never going to be exactly $0$ or~$1$,
but it is nonetheless a useful quantity when performing approximate inference,
as we will see in~\cref{sec:rel}.

\subsection{Variational Inference}
\label{sec:var}

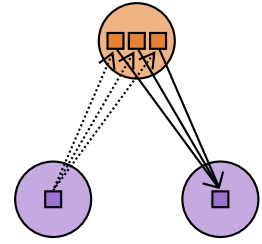
\begin{figure}
\centering

\begin{tikzpicture}[default]

\node[ent] (x) {$x$} ;
\node[val, below=of x, xshift=-11mm] (t1) {} ;
\node[val, below=of x, xshift=11mm] (t2) {} ;

\node[comp] (x1) at (x) [xshift=-\compsep] {} ;
\node[comp] (x2) at (x) {};
\node[comp] (x3) at (x) [xshift=\compsep] {};

\node[sig] (s1) at (t1) {} ;
\node[sig] (s2) at (t2) {} ;

\draw (s2.north) edge[dirextra] (x1.-80) edge[dircomp] (x2.-80) edge[dirextra] (x3.-80) ;

\foreach \i in {1,2,3} {
	\draw (x\i.-100) edge[inf] (s1.north) ;
}

\end{tikzpicture}
\vspace*{-3mm}

\caption{Variational inference for \cref{fig:log-one}.
Exactly calculating the distribution of $x$ given~$t_{a,x}$ is intractable,
but we can use a mean-field approximation.
The dotted lines indicate approximate inference,
and the solid lines indicate inference from the mean-field pixie vector.
}
\label{fig:var-one}

\end{figure}

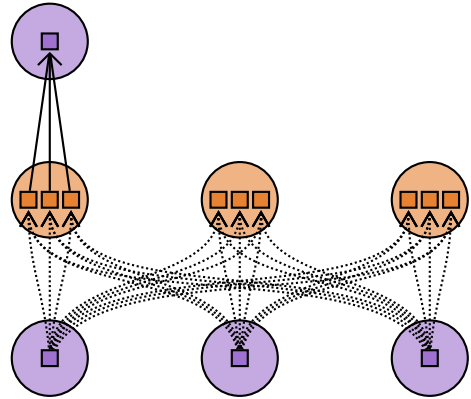
\begin{figure}
\centering
\begin{tikzpicture}[default]


\node[ent] (y) {} ;
\node[ent, right=of y] (z) {} ;
\node[ent, left=of y] (x) {} ;


\node[comp] (x1) at (x) [xshift=-\compsep] {};
\node[comp] (x2) at (x) {};
\node[comp] (x3) at (x) [xshift=\compsep] {};
\node[comp] (y1) at (y) [xshift=-\compsep] {};
\node[comp] (y2) at (y) {};
\node[comp] (y3) at (y) [xshift=\compsep] {};
\node[comp] (z1) at (z) [xshift=-\compsep] {};
\node[comp] (z2) at (z) {};
\node[comp] (z3) at (z) [xshift=\compsep] {};


\node[val, below=of x] (tx) {} ;
\node[val, below=of y] (ty) {} ;
\node[val, below=of z] (tz) {} ;
\node[val, above=of x] (t2) {} ;


\node[sig] (sx) at (tx) {} ;
\node[sig] (sy) at (ty) {} ;
\node[sig] (sz) at (tz) {} ;
\node[sig] (s2) at (t2) {} ;

\draw (s2.south) edge[dirextra] (x1) edge[dircomp] (x2) edge[dirextra] (x3);


\foreach \x in {x,y,z} {
\foreach \i in {1,2,3} {
  \draw (\x\i) edge[inf, in=100-5*\i, out=-90] (s\x);
}}

\foreach \x/\y in {x/y, y/z} {
\foreach \i in {1,2,3} {
  \draw (\x\i) edge[inf, in=110-5*\i, out=-90, looseness=0.9] (s\y);
}}

\foreach \x/\y in {z/y, y/x} {
\foreach \i in {1,2,3} {
  \draw (\x\i) edge[inf, in=90-5*\i, out=-90, looseness=0.9] (s\y);
}}

\foreach \i in {1,2,3} {
  \draw (x\i) edge[inf, in=126-7*\i, out=-90, looseness=0.7] (sz);
}

\foreach \i in {1,2,3} {
  \draw (z\i) edge[inf, in=82-7*\i, out=-90, looseness=0.7] (sx);
}

\end{tikzpicture}
\vspace*{-3mm}

\caption{Variational inference for \cref{fig:log-three}.
The \textsc{arg1} and \textsc{arg2} links are not explicitly represented,
but the mean-field probabilities are optimised
to approximate the joint distribution induced by the links.
Each dimension has an independent probability,
but they are jointly optimised,
so they depend on all truth values for all pixies.
}
\label{fig:var-three}
\vspace*{-3mm}

\end{figure}

As explained in the previous section,
we can express certain logical propositions as conditional probabilities,
but calculating these probabilities exactly is intractable,
as it involves summing over the semantic space,
which grows exponentially with the number of dimensions.
Furthermore, we need to calculate similar conditional probabilities
when training the model in the first place.

Instead of summing over the entire space, E\&C proposed
summing over a small number of carefully chosen pixies,
using a Markov Chain Monte Carlo method.
However, this algorithm is slow for two reasons.
Firstly, many iterations of the Markov chain are required
before the samples are useful.
Secondly, even if we are not summing over the entire space,
many samples are still needed,
because the discrete values lead to high variance.

In this section, we introduce a variational inference algorithm,
where we directly approximate the distribution over pixies that we need to calculate,
and then optimise this approximation.
This makes the calculations in the previous section tractable,
and also makes training more efficient.

The distribution we need to approximate is ${P(x|t_{c,x})}$,
the probability that a latent pixie~$x$ has particular features,
given the truth of some predicate~$c$.
We use a mean-field approximation:
we assume that each dimension has an independent probability~$q_i$ of being active,
as shown in~\cref{eqn:var-approx}.
The approximate probability~$Q(x)$ is simply the product of the probabilities of each dimension.
Furthermore, we assume that each of these probabilities
depends on the average activation of all other dimensions
(i.e.\ the \textit{mean field} activation).
\begin{equation}
P(x|t_{c,x}) \approx Q(x) = \!\prod_{i|x_i=1}\! q_i \!\prod_{i|x_i=0}\! (1-q_i)
\label{eqn:var-approx}
\end{equation}

For~$Q$ to be a good approximation, it needs to be close to~$P$.
We can measure this using the Kullback-Leibler divergence from $Q$~to~$P$.\footnote{%
	Variational Bayes minimises the KL-divergence in the opposite direction
	-- that is, the KL-divergence from $P$~to~$Q$.
	However, for the above approximation, this is infinite:
	if the number of active units is not equal to the fixed cardinality,
	then $P(x|t_{c,x})=0$ but $Q(x) \neq 0$,
	giving infinite $Q(x)\log P(x|t_{c,x})$.
	Furthermore, while Variational Bayes prefers `high precision' approximations
	(areas of high~$Q$ are accurate),
	we will prefer `high recall' approximations
	(areas of high~$P$ are accurate).
	This is appropriate for two reasons.
	Firstly, in areas where the number of active units is wrong,
	$Q$ is bound to be too high, but if we want to sample from~$Q$,
	we can avoid these areas by using belief propagation,
	as explained by \newcite{swersky2012carbm}.
	Secondly, in areas where the number of active units is correct,
	$Q$ will be much higher than~$P$ only if there is a dependence between dimensions that $Q$ cannot capture,
	such as if $P$ is a multi-modal distribution.
	Because of the definition of an RBM, such a dependence is impossible within one pixie,
	and combined with the simple form of our semantic functions,
	such a dependence will be rare between pixies.
}
Minimising this quantity is also done in the Expectation Propagation algorithm \cite{minka2001expectation}.
However, a semantic function model is not in the exponential family,
making it difficult to apply Expectation Propagation directly.

Given this mean-field approximation~$Q(x)$,
we have a a mean-field vector~$q_i$.
This vector is not in~$\mathcal{X}$,
because each dimension is now a value in the range~$[0,1]$,
rather than taking one of the values $0$ or~$1$.
It represents a `typical' pixie for these truth values.
Furthermore, we have implemented semantic functions as one-layer neural networks,
and each weight in the network can be multiplied by a value in the range~$[0,1]$
just as easily as it can be multiplied by $0$ or~$1$.
Since a mean-field vector defines a distribution over pixies,
applying a semantic function to a mean-field vector
lets us approximately calculate the probability that
a predicate is true of a pixie drawn from this distribution.

Differentiating the KL-divergence with respect to~$q_i$,
and using the above idea that we can apply semantic functions to mean-field vectors,
we can derive the update rule given in~\cref{eqn:update},
with a full derivation given in \cref{append:update}.
This updates the value of~$q_i$, while holding the rest fixed.
Here, $x^{(+i)}$ is the mean-field vector where unit~$i$ is fixed to be on,
$x^{(-i)}$ is the mean-field vector where unit~$i$ is fixed to be off,
$t_c$ is the semantic function for the predicate~$c$,
$D$ is the number of dimensions of~$\mathcal{X}$,
and $C$ is the number of active units.
Optimising $Q$ can then be done by repeatedly applying this update rule across all dimensions.
\vspace*{-1mm}
\begin{equation}
q_i = \left(1 + \frac{D-C}{C} \frac{t_c\left(x^{(-i)}\right)}{t_c\left(x^{(+i)}\right)} \right)^{-1}
\label{eqn:update}
\end{equation}

This update rule looks at how likely the predicate~$c$ is to be true
when the dimension~$x_i$ is active, and when it is not.
If $c$ is much more likely to be true when $x_i$ is active,
then $q_i$ will be close to~$1$.
If $c$ is much more likely to be true when $x_i$ is inactive,
then $q_i$ will be close to~$0$.
If there is no difference at all,
then $q_i$ will be $C/D$, the expected probability if all dimensions are equally likely.

We can apply this to logical inference,
to calculate $P(t_{b,x}|t_{a,x})$, as shown in \cref{fig:var-one}.
We first find the mean-field vector for~$x$, conditioning on the truth of~$a$.
This approximates $P(x|t_{a,x})$.
Then, we evaluate the semantic function for $b$ on this mean-field vector.
This approximates $P(t_{b,x}|t_{a,x})$.

For multiple pixies, the process is similar,
as shown in \cref{fig:var-three}.
We have one mean-field vector for each pixie,
and we optimise these together.
The only difference to the update rule is that, as well as considering
how activating one dimension changes the probability of a predicate being true,
we also have to consider how likely this dimension is to be active,
given the other pixies in the graph.
This leads to an extra term in the update rule, as exemplified in~\cref{eqn:update-multi},
where there is a link from $x$ to~$y$.
The link has weights~$W_{ij}$ which control how likely it is that $x_i$ and~$y_j$ are both active.
\vspace*{-3mm}
\begin{equation}
q_i = \left(1 + \frac{D-C}{C}
  \frac{t_c\left(x^{(-i)}\right)}{t_c\left(x^{(+i)}\right)}
  e^{-\Sigma_j W_{ij}y_j}
\right)^{-1}
\label{eqn:update-multi}
\end{equation}

\begin{table*}[h]
\center
\begin{tabularx}{.91\textwidth}{|l|C|C|C|C|}

\hline
Model & SimLex Nouns & SimLex Verbs & WordSim Sim. & WordSim Rel. \\ \hline
Word2Vec & .40 & \bf .23 & \bf .69 & .46 \\
SVO Word2Vec & .44 & .18 & .61 & .24 \\
Semantic Functions & \bf .46 & .19 & .60 & \bf .14 \\ \hline

\end{tabularx}
\vspace*{-1mm}
\caption{\centering
Spearman rank correlation with average annotator judgements.
Note that we would like\newline to have a \emph{low} score on the final column
(which measures relatedness, rather than similarity).
}
\label{tab:sim}
\end{table*}

\begin{table*}[h]
\center
\begin{tabularx}{.68\textwidth}{|l|C|C|}

\hline
Model & Development & Test \\ \hline
Word2Vec, Addition & .50 & .47 \\
Semantic Functions & .20 & .16 \\
Word2Vec and Sem-Func Ensemble & \bf .53 & \bf .49 \\ \hline

\end{tabularx}
\vspace*{-1mm}
\caption{\centering
Mean average precision on the RELPRON development and test sets.
Note that this Word2Vec model was trained on
a more recent (and hence larger) version of Wikipedia,
to match \citeauthor{rimell2016relpron}
}
\label{tab:rel}
\vspace*{-3mm}
\end{table*}

\section{Experimental Results\footnote{%
	A fuller set of results, with further discussion,
	will be given by E\&C-forth.
}}

Finding a good evaluation task is difficult.
Lexical similarity tasks do not require logical inference,
while tasks like textual entailment
require a level of coverage beyond the scope of this paper.
We consider two tasks:
lexical similarity, as a simple benchmark,
and the RELPRON dataset, which lets us explore a controlled kind of inference.

We trained our model on subject-verb-object (SVO) triples extracted from
WikiWoods\footnote{\url{http://moin.delph-in.net/WikiWoods}},
a parsed version of the July 2008 dump of the English Wikipedia,
distributed by DELPH-IN.
This resource was produced by \newcite{flickinger2010wikiwoods},
using the English Resource Grammar \cite{flickinger2000erg,flickinger2011erg},
and the PET parser \cite{callmeier2001pet,toutanova2005pet},
with parse ranking trained on the manually treebanked subcorpus WeScience \cite{ytrestol2009wescience}.

Our source code is available online.\footnote{\url{http://github.com/guyemerson/sem-func}}
The WikiWoods corpus was pre-processed using the Python packages
pydelphin\footnote{\url{http://github.com/delph-in/pydelphin}} (developed by Michael Goodman),
and pydmrs\footnote{\url{http://github.com/delph-in/pydmrs}} \cite{copestake2016pydmrs}.

To speed up training, we initialised our model using random positive-only projections,
a simple method for producing reduced-dimension count vectors \cite{qasemizadeh2016vector}.
Rather than counting each context separately,
every context is randomly mapped to a dimension,
so each dimension corresponds to the total count of several contexts.
These counts can then be transformed into PPMI scores.
As with normal PPMI-based count vectors,
there are several hyperparameters that can be tuned \cite{levy2015hyperparam} --
however, as we are using these vectors as parameters for semantic functions,
it should be noted that the optimal hyperparameter settings are not the same.

We compare our model to two vector space models, also trained on Wikipedia.
Both use \newcite{mikolov2013vector}'s
skipgram algorithm with negative sampling.
 ``Word2Vec'' was trained on raw text, while
``SVO Word2Vec'' was trained on the same SVO triples used to train our model.
We tested these models using cosine similarity.

\subsection{Lexical Semantic Similarity}
\label{sec:sim}

To measure the similarity of two predicates $a$ and~$b$,
we use the conditional probability described in~\cref{sec:log},
and illustrated in \cref{fig:log-one,fig:var-one}.
Since this is an asymmetric measure,
we multiply the conditional probabilities in both directions,
i.e.\ we calculate ${P(t_{a,x}|t_{b,x})P(t_{b,x}|t_{a,x})}$.

We evaluated on two datasets which aim to capture similarity, rather than relatedness:
SimLex-999 \cite{hill2015simlex},
and WordSim-353 \cite{finkelstein2001wordsim}, 
which \newcite{agirre2009wordsim} split into similarity and relatedness subsets.
Results are shown in \cref{tab:sim}.\footnote{%
  Performance of Word2Vec on SimLex-999 is higher than reported by \newcite{hill2015simlex}.
  Despite correspondence with the authors,
  it is not clear why their figures are so low.
}
For each dataset, hyperparameters were tuned on the remaining datasets.
As WordSim-353 is a noun-based dataset,
it is possible that performance on SimLex-999 verbs could be improved
by optimising hyperparameters on a more appropriate development set.

Note that we would like a \textit{low} correlation on the relatedness subset of WordSim-353.
In the real world, related predicates are unlikely to be true of the same pixies
(and the pixies they are true of are unlikely to even share features).
For predicates which are true of similar but disjoint sets of pixies,
annotations in these datasets are inconsistent.
For example, SimLex-999 gives a low score to (\textit{aunt}, \textit{uncle}),
but a high score to (\textit{cat}, \textit{dog}).
The semantic function model achieves the lowest correlation on the relatedness subset.

Compared to E\&C,
the gap between the semantic function model and the vector space models has essentially been closed.
Which model performs best is inconsistent across the evaluation datasets.
This shows that the previously reported lower performance was not due to a problem with the model itself,
but rather with an inefficient training algorithm and with poor choice of hyperparameters.

\subsection{RELPRON}
\label{sec:rel}

The RELPRON dataset was produced by \newcite{rimell2016relpron}.
It consists of `terms' (all nouns), each paired with up to ten `properties'.
For example,
a \textit{telescope} is a \textit{device that astronomers use},
and a \textit{saw} is a \textit{device that cuts wood}.
All properties are of this form:
a hypernym of the term,
modified by a relative clause with a transitive verb.
For each term,
the task is to identify the properties which apply to this term.
Since every property follows one of only two patterns,
this dataset lets us focus on semantics, rather than parsing.

A model that captures relatedness can achieve good performance on this dataset --
\citeauthor{rimell2016relpron} found that the \textit{other} argument of the verb
was the best predictor of the term
(e.g.\ \textit{astronomer} predicts \textit{telescope}).
Logically speaking, these predicates do not imply each other.
However, \citeauthor{rimell2016relpron} included confounders for a model relying on relatedness --
e.g.\ a \textit{document that has a balance} is an \textit{account},
not the quality of \textit{balance}.
In all of their models, this was the top-ranked property for \textit{balance}.
By combining a vector model with our model,
we hoped to improve performance.

We tested our model using the method described in~\cref{sec:theory}
and illustrated in \cref{fig:log-three,fig:var-three}:
for each term and property, we find the probability of the term being true,
conditioned on all predicates in the property.
Results are given in \cref{tab:rel}.
As noted in~\cref{sec:sim}, our model does not capture relatedness,
and it performs below vector addition.
However, the ensemble outperforms the vector space model alone.
This improvement is not simply due to increasing the capacity of the model --
increasing the dimensionality of the vector space did not yield this improvement.

Furthermore, inspecting the differences in predictions
between the vector space model and the ensemble,
it appears that there is particular improvement on
the confounders included in the dataset,
which require some kind of logical inference.
In our ensemble model, for the term \textit{balance},
the top-ranked property is no longer the confounder \textit{document that has a balance},
but instead the correct property \textit{quality that an ear maintains}.


\section{Conclusion}

We can define probabilistic logical inference in Functional Distributional Semantics,
and efficiently calculate it using variational inference.
We can use this to improve performance on the RELPRON dataset,
suggesting our model can learn structure
not captured by vector space models.

\section*{Acknowledgements}

We would like to thank Emily Bender,
for helpful discussion and feedback on an earlier draft.

This work was supported by a Schiff Foundation Studentship.

\bibliography{emerson-laml}

\begin{thebibliography}{}
\expandafter\ifx\csname natexlab\endcsname\relax\def\natexlab#1{#1}\fi

\bibitem[{Agirre et~al.(2009)Agirre, Alfonseca, Hall, Kravalova, Pa{\c{s}}ca,
  and Soroa}]{agirre2009wordsim}
Eneko Agirre, Enrique Alfonseca, Keith Hall, Jana Kravalova, Marius
  Pa{\c{s}}ca, and Aitor Soroa. 2009.
\newblock A study on similarity and relatedness using distributional and
  wordnet-based approaches.
\newblock In {\em Proceedings of the 2009 Conference of the North American
  Chapter of the Association for Computational Linguistics\/}. pages 19--27.

\bibitem[{Baroni et~al.(2014)Baroni, Bernardi, and
  Zamparelli}]{baroni2014tensor}
Marco Baroni, Raffaela Bernardi, and Roberto Zamparelli. 2014.
\newblock Frege in space: A program of compositional distributional semantics.
\newblock {\em Linguistic Issues in Language Technology\/} 9.

\bibitem[{Beltagy et~al.(2016)Beltagy, Roller, Cheng, Erk, and
  Mooney}]{beltagy2016logic}
Islam Beltagy, Stephen Roller, Pengxiang Cheng, Katrin Erk, and Raymond~J
  Mooney. 2016.
\newblock Representing meaning with a combination of logical and distributional
  models.
\newblock {\em Computational Linguistics\/} 42(4):763--808.

\bibitem[{Bergmair(2010)}]{bergmair2010fuzzy}
Richard Bergmair. 2010.
\newblock {\em Monte Carlo Semantics: Robust inference and logical pattern
  processing with Natural Language text\/}.
\newblock Ph.D. thesis, University of Cambridge.

\bibitem[{Callmeier(2001)}]{callmeier2001pet}
Ulrich Callmeier. 2001.
\newblock {\em Efficient parsing with large-scale unification grammars\/}.
\newblock Master's thesis, Universit{\"a}t des Saarlandes, Saarbr{\"u}cken,
  Germany.

\bibitem[{Clarke and Keller(2015)}]{clarke2015stochastic}
Daoud Clarke and Bill Keller. 2015.
\newblock Efficiency in ambiguity: Two models of probabilistic semantics for
  natural language.
\newblock In {\em Proceedings of the 11th International Conference on
  Computational Semantics (IWCS)\/}. pages 129--139.

\bibitem[{Coecke et~al.(2010)Coecke, Sadrzadeh, and Clark}]{coecke2010tensor}
Bob Coecke, Mehrnoosh Sadrzadeh, and Stephen Clark. 2010.
\newblock Mathematical foundations for a compositional distributional model of
  meaning.
\newblock {\em Linguistic Analysis\/} 36:345--384.

\bibitem[{Cooper(2005)}]{cooper2005type}
Robin Cooper. 2005.
\newblock Austinian truth, attitudes and type theory.
\newblock {\em Research on Language and Computation\/} 3(2-3):333--362.

\bibitem[{Cooper et~al.(2015)Cooper, Dobnik, Larsson, and
  Lappin}]{cooper2015prob}
Robin Cooper, Simon Dobnik, Staffan Larsson, and Shalom Lappin. 2015.
\newblock Probabilistic type theory and natural language semantics.
\newblock {\em LiLT (Linguistic Issues in Language Technology)\/} 10.

\bibitem[{Copestake(2009)}]{copestake2009dmrs}
Ann Copestake. 2009.
\newblock Slacker semantics: Why superficiality, dependency and avoidance of
  commitment can be the right way to go.
\newblock In {\em Proceedings of the 12th Conference of the European Chapter of
  the Association for Computational Linguistics\/}. pages 1--9.

\bibitem[{Copestake et~al.(2016)Copestake, Emerson, Goodman, Horvat, Kuhnle,
  and Muszy{\'n}ska}]{copestake2016pydmrs}
Ann Copestake, Guy Emerson, Michael~Wayne Goodman, Matic Horvat, Alexander
  Kuhnle, and Ewa Muszy{\'n}ska. 2016.
\newblock Resources for building applications with {D}ependency {M}inimal
  {R}ecursion {S}emantics.
\newblock In {\em Proceedings of the 10th International Conference on Language
  Resources and Evaluation (LREC)\/}. European Language Resources Association
  (ELRA).

\bibitem[{Copestake et~al.(2005)Copestake, Flickinger, Pollard, and
  Sag}]{copestake2005mrs}
Ann Copestake, Dan Flickinger, Carl Pollard, and Ivan~A Sag. 2005.
\newblock Minimal {R}ecursion {S}emantics: An introduction.
\newblock {\em Research on Language and Computation\/} 3(2-3):281--332.

\bibitem[{Copestake and Herbelot(2012)}]{copestake2012ideal}
Ann Copestake and Aur{\'e}lie Herbelot. 2012.
\newblock Lexicalised compositionality.
\newblock Unpublished draft.

\bibitem[{Emerson and Copestake(2016)}]{emerson2016}
Guy Emerson and Ann Copestake. 2016.
\newblock Functional {D}istributional {S}emantics.
\newblock In {\em Proceedings of the 1st Workshop on Representation Learning
  for NLP (RepL4NLP)\/}. Association for Computational Linguistics, pages
  40--52.

\bibitem[{Emerson and Copestake(2017)}]{emerson-forth}
Guy Emerson and Ann Copestake. 2017.
\newblock Semantic composition via probabilistic model theory.
\newblock In {\em Proceedings of the 12th International Conference on
  Computational Semantics (IWCS)\/}. Association for Computational Linguistics.

\bibitem[{Finkelstein et~al.(2001)Finkelstein, Gabrilovich, Matias, Rivlin,
  Solan, Wolfman, and Ruppin}]{finkelstein2001wordsim}
Lev Finkelstein, Evgeniy Gabrilovich, Yossi Matias, Ehud Rivlin, Zach Solan,
  Gadi Wolfman, and Eytan Ruppin. 2001.
\newblock Placing search in context: The concept revisited.
\newblock In {\em Proceedings of the 10th International Conference on the World
  Wide Web\/}. Association for Computing Machinery, pages 406--414.

\bibitem[{Flickinger(2000)}]{flickinger2000erg}
Dan Flickinger. 2000.
\newblock On building a more efficient grammar by exploiting types.
\newblock {\em Natural Language Engineering\/} 6(1):15--28.

\bibitem[{Flickinger(2011)}]{flickinger2011erg}
Dan Flickinger. 2011.
\newblock Accuracy vs. robustness in grammar engineering.
\newblock In Emily~M Bender and Jennifer~E Arnold, editors, {\em Language from
  a cognitive perspective: Grammar, usage, and processing\/}, CSLI
  Publications, pages 31--50.

\bibitem[{Flickinger et~al.(2010)Flickinger, Oepen, and
  Ytrest{\o}l}]{flickinger2010wikiwoods}
Dan Flickinger, Stephan Oepen, and Gisle Ytrest{\o}l. 2010.
\newblock Wiki{W}oods: Syntacto-semantic annotation for {E}nglish {W}ikipedia.
\newblock In {\em Proceedings of the 7th International Conference on Language
  Resources and Evaluation (LREC)\/}. European Language Resources Association
  (ELRA).

\bibitem[{Garrette et~al.(2011)Garrette, Erk, and Mooney}]{garrette2011logic}
Dan Garrette, Katrin Erk, and Raymond Mooney. 2011.
\newblock Integrating logical representations with probabilistic information
  using {M}arkov logic.
\newblock In {\em Proceedings of the 9th International Conference on
  Computational Semantics (IWCS)\/}. Association for Computational Linguistics,
  pages 105--114.

\bibitem[{Grefenstette(2013)}]{grefenstette2013tensor}
Edward Grefenstette. 2013.
\newblock Towards a formal distributional semantics: Simulating logical calculi
  with tensors.
\newblock In {\em Proceedings of the 2nd Joint Conference on Lexical and
  Computational Semantics (*SEM)\/}. pages 1--10.

\bibitem[{Hill et~al.(2015)Hill, Reichart, and Korhonen}]{hill2015simlex}
Felix Hill, Roi Reichart, and Anna Korhonen. 2015.
\newblock Simlex-999: Evaluating semantic models with (genuine) similarity
  estimation.
\newblock {\em Computational Linguistics\/} 41(4):665--695.

\bibitem[{Larsson(2013)}]{larsson2013classifier}
Staffan Larsson. 2013.
\newblock Formal semantics for perceptual classification.
\newblock {\em Journal of Logic and Computation\/} 25(2):335--369.

\bibitem[{Levy et~al.(2015)Levy, Goldberg, and Dagan}]{levy2015hyperparam}
Omer Levy, Yoav Goldberg, and Ido Dagan. 2015.
\newblock Improving distributional similarity with lessons learned from word
  embeddings.
\newblock {\em Transactions of the Association for Computational Linguistics
  (TACL)\/} 3:211--225.

\bibitem[{Mikolov et~al.(2013)Mikolov, Chen, Corrado, and
  Dean}]{mikolov2013vector}
Tomas Mikolov, Kai Chen, Greg Corrado, and Jeffrey Dean. 2013.
\newblock Efficient estimation of word representations in vector space.
\newblock In {\em Proceedings of the 1st International Conference on Learning
  Representations\/}.

\bibitem[{Minka(2001)}]{minka2001expectation}
Thomas~P Minka. 2001.
\newblock Expectation propagation for approximate {B}ayesian inference.
\newblock In {\em Proceedings of the 17th Conference on Uncertainty in
  Artificial Intelligence\/}. pages 362--369.

\bibitem[{QasemiZadeh and Kallmeyer(2016)}]{qasemizadeh2016vector}
Behrang QasemiZadeh and Laura Kallmeyer. 2016.
\newblock Random positive-only projections: {PPMI}-enabled incremental semantic
  space construction.
\newblock In {\em Proceedings of the 5th Joint Conference on Lexical and
  Computational Semantics (*SEM)\/}. pages 189--198.

\bibitem[{Rimell et~al.(2016)Rimell, Maillard, Polajnar, and
  Clark}]{rimell2016relpron}
Laura Rimell, Jean Maillard, Tamara Polajnar, and Stephen Clark. 2016.
\newblock {RELPRON}: A relative clause evaluation dataset for compositional
  distributional semantics.
\newblock {\em Computational Linguistics\/} 42(4):661--701.

\bibitem[{Schlangen et~al.(2016)Schlangen, Zarrie{\ss}, and
  Kennington}]{schlangen2016classifier}
David Schlangen, Sina Zarrie{\ss}, and Casey Kennington. 2016.
\newblock Resolving references to objects in photographs using the
  words-as-classifiers model.
\newblock In {\em The 54th Annual Meeting of the Association for Computational
  Linguistics\/}. pages 1213--1223.

\bibitem[{Swersky et~al.(2012)Swersky, Sutskever, Tarlow, Zemel, Salakhutdinov,
  and Adams}]{swersky2012carbm}
Kevin Swersky, Ilya Sutskever, Daniel Tarlow, Richard~S Zemel, Ruslan~R
  Salakhutdinov, and Ryan~P Adams. 2012.
\newblock Cardinality {R}estricted {B}oltzmann {M}achines.
\newblock In {\em Advances in Neural Information Processing Systems 25
  (NIPS)\/}. pages 3293--3301.

\bibitem[{Toutanova et~al.(2005)Toutanova, Manning, and
  Oepen}]{toutanova2005pet}
Kristina Toutanova, Christopher~D Manning, and Stephan Oepen. 2005.
\newblock Stochastic {HPSG} parse selection using the {R}edwoods corpus.
\newblock {\em Journal of Research on Language and Computation\/} 3(1):83--105.

\bibitem[{Turney and Pantel(2010)}]{turney2010vector}
Peter~D. Turney and Patrick Pantel. 2010.
\newblock From frequency to meaning: Vector space models of semantics.
\newblock {\em Journal of Artificial Intelligence Research\/} 37:141--188.

\bibitem[{Ytrest{\o}l et~al.(2009)Ytrest{\o}l, , Oepen, and
  Flickinger}]{ytrestol2009wescience}
Gisle Ytrest{\o}l, , Stephan Oepen, and Dan Flickinger. 2009.
\newblock Extracting and annotating {W}ikipedia sub-domains.
\newblock In {\em Proceedings of the 7th International Workshop on Treebanks
  and Linguistic Theories\/}.

\bibitem[{Zarrie{\ss} and Schlangen(2017)}]{zarriess2017classifier}
Sina Zarrie{\ss} and David Schlangen. 2017.
\newblock Is this a child, a girl or a car? {E}xploring the contribution of
  distributional similarity to learning referential word meanings.
\newblock In {\em Proceedings of the 15th Annual Conference of the European
  Chapter of the Association for Computational Linguistics\/}. pages 86--91.

\end{thebibliography}
\bibliographystyle{acl_natbib}

\vspace*{0mm}

\appendix

\section{Logical equivalence}
\label{append:equiv}

\subsection{Proof of general case}

Syllogisms are classically expressed in set-theoretic terms.
A quantified proposition of the form \textit{Q A are B},
where \textit{Q} is some quantifier,
gives constraints on the sizes of the sets ${A\cap B}$ and~${A\setminus B}$,
and says nothing about the size of~$B$.

For the quantifier~$\exists$, we have:
\begin{equation*}
{|A \cap B | > 0}
\end{equation*}

For the quantifier~$\forall$, we have the following,
where the second constraint assumes existential import:
\begin{align*}
|A\setminus B| & = 0 \\
|A\cap B| & > 0
\end{align*}

From these definitions, we can use standard set theory
to prove all and only the valid syllogisms.
To show equivalence with our probabilistic framework,
we first note that sizes of sets form a measure (the `counting measure'),
and probabilities also form a measure.
The above conditions are all constraints on sizes of sets being zero or nonzero,
so it suffices to show that the sizes and probabilities are equivalent in the measure-theoretic sense:
they agree on which sets have measure zero.

First, we note that ${P(B|A)=\frac{P(A\cap B)}{P(A)}}$ is defined only when $P(A)>0$,
which will give us existential import.

For~$\exists$, we have:
\begin{align*}
P(B|A)=\frac{P(A\cap B)}{P(A)} & > 0 \\
P(A\cap B) & > 0
\end{align*}

We can say nothing further about the probability ${P(A\setminus B) = P(A) - P(A\cap B)}$,
which may be zero or nonzero,
just as in the classical case.

For~$\forall$, we have:
\begin{align*}
\frac{P(A\cap B)}{P(A)} & = 1 \\
P(A\cap B) & = P(A) \\
P(A\cap B) & = P(A\cap B) + P(A\setminus B) \\
P(A\setminus B) & = 0
\end{align*}

And we also have:
\begin{align*}
P(A\setminus B) + P(A\cap B) & = P(A) > 0 \\
P(A\cap B) & > 0
\end{align*}

This demonstrates the equivalence.

\subsection{Example}

We can prove the Barbara syllogism as follows:
\begin{align*}
P(B|A) = 1 \implies& P(A\setminus B) = 0, \\
& P(A) > 0 \\
P(C|B) = 1 \implies& P(B\setminus C) = 0
\end{align*}
\begin{align*}
P(A\setminus C) &= P(A\cap B\setminus C) + P(A\setminus B\setminus C) \\
&\leq P(B\setminus C) + P(A\setminus B) \\
&= 0
\end{align*}
\begin{align*}
P(A\cap C) &= P(A) - P(A\setminus C) \\
&= P(A) > 0
\end{align*}
\begin{equation*}
\implies P(C|A) = \frac{P(A\cap C)}{P(A)} = 1
\end{equation*}

\pagebreak

\section{Derivation of update rule}
\label{append:update}

\newcommand{\dq}{\frac{\partial}{\partial q_i}}

We are trying to optimise $Q$ to minimise the KL-divergence from $Q(x)$ to~$P(x|t_{c,x})$:
\begin{equation*}
\begin{split}
D_{\mathrm{KL}}(P||Q) = \sum_x P(x|t_{c,x}) \, \log\frac{P(x|t_{c,x})}{Q(x)} \\
= \sum_x P(x|t_{c,x}) \big( \log P(x|t_{c,x}) - \log Q(x) \big)
\end{split}
\end{equation*}

Note that the first term is independent of~$Q$.
To iteratively optimise one parameter~$q_i$ at a time,
we take the derivative:
\begin{equation*}
\begin{split}
\dq D_{\mathrm{KL}}(P||Q) = -\dq \sum_x P(x|t_{c,x}) \log Q(x) \\
= \sum_{x | x_i = 1} P(x|t_{c,x}) \frac{1}{q_i} - \sum_{x | x_i = 0} P(x|t_{c,x}) \frac{1}{1 - q_i}
\end{split}
\end{equation*}

Now we can rewrite $P(x|t_{c,x})$ as the following.
If there is just one pixie, then we can assume a uniform prior over~$x$.
For $D$~dimensions, of which $C$ are active, there are $D \choose C$ different vectors.
\begin{align*}
P(x|t_{c,x}) & = \frac{P(x)P(t_{c,x}|x)}{P(t_{c,x})} \\
& = \frac{t_c(x)}{{D \choose C}P(t_{c,x})}
\end{align*}

Note that ${D \choose C}P(t_{c,x})$ is constant in~$x$.
Setting the derivative to~$0$, we have:
\begin{equation*}
\sum_{x | x_i = 1} t_c(x) \frac{1}{q_i} = \sum_{x | x_i = 0} t_c(x) \frac{1}{1 - q_i}
\end{equation*}

Summing over all~$x$ is intractable,
but we can approximate this sum using mean-field vectors for~$x$.
For most values of~$x$, $t_c(x)$ will be close to~$0$,
and the regions of interest will be near the mean-field vectors.
Let $x^{(+i)}$ denote the mean-field vector
when $x_i=1$ and the total activation of the remaining dimensions is~$C-1$,
and let $x^{(-i)}$ denote the mean-field vector
when $x_i=0$ and the total activation of the remaining dimensions is~$C$.
Both of these vectors can be approximated using the values of~$q_j$ for $j \neq i$,
scaled so that their sum is correct.
Then we have:
\begin{align*}
{D\!-\!1 \choose C\!-\!1} t_c(x^{(+i)}) \frac{1}{q_i} & \approx {D\!-\!1 \choose C} t_c(x^{(-i)}) \frac{1}{1 - q_i} \\
t_c(x^{(+i)}) \frac{1}{q_i} & \approx \frac{D\!-\!C}{C} t_c(x^{(-i)}) \frac{1}{1 - q_i}
\end{align*}

Re-arranging for $q_i$ yields the following,
which is the optimal value for~$q_i$,
given the other dimensions~$q_j$,
and given the above approximations:
\begin{equation*}
q_i \approx \left(1 + \frac{D-C}{C} \frac{t_c\left(x^{(-i)}\right)}{t_c\left(x^{(+i)}\right)} \right)^{-1}
\end{equation*}

In the above derivation,
we assumed a uniform prior over~$x$,
which meant that $P(x|t_{c,x}) \propto t_c(x)$.
If there are links between pixies,
then this no longer holds,
and we instead have $P(x)$
being determined by the RBM weights, which gives the following,
where we sum over all links $x \xrightarrow{l} y$,
from the pixie~$x$ to another pixie~$y$ with label~$l$.
Each link type~$l$ has weights~$W^{(l)}_{jk}$
(and for incoming links, we simply take the transpose of this matrix).
For clarity, we do not write bias terms.
\begin{equation*}
P(x|t_{c,x}) \propto t_c(x) \exp \!\sum_{x \, \xrightarrow{l} \, y}\! \sum_{j,k} W^{(l)}_{jk}x_j y_k
\end{equation*}

So to amend the update rule,
we replace $t_c(x)$ with the above expression, which gives:
\begin{equation*}
\left(1 + \frac{D-C}{C} \frac
{t_c\left(x^{(-i)}\right) \exp\sum W^{(l)}_{jk}x^{(-i)}_j y_k}
{t_c\left(x^{(+i)}\right) \exp\sum W^{(l)}_{jk}x^{(+i)}_j y_k}
\right)^{-1}
\end{equation*}

Now note that this ratio of exponentials can be rewritten as:
\begin{equation*}
\exp\sum W^{(l)}_{jk}\left(x^{(-i)}_j - x^{(+i)}_j\right) y_k
\end{equation*}

For dimensions $j \neq i$, the difference between the two mean-field vectors will be small,
so if $\sum_k W^{(l)}_{jk} y_k$ is on average close to~$0$,
the above expression will be dominated by the value at $j=i$.
So, we can approximate it as:
\begin{equation*}
\exp - \!\!\sum_{x \, \xrightarrow{l} \, y}\! \sum_k W^{(l)}_{ik} y_k
\end{equation*}

This gives the following update rule, which
reduces to \cref{eqn:update-multi} in the case of a single link:
\begin{equation*}
\left(1 + \frac{D\!-\!C}{C} \frac{t_c\left(x^{(-i)}\right)}{t_c\left(x^{(+i)}\right)}
\exp - \!\!\! \sum_{x \, \xrightarrow{l} \, y} \!\!\sum_k W^{(l)}_{ik} y_k
\right)^{\!\!-1}
\end{equation*}

\end{document}